\documentclass{article}

\usepackage{microtype}
\usepackage{graphicx}
\usepackage{subfigure}
\usepackage{booktabs} 
\usepackage{natbib}

\usepackage{hyperref}

\usepackage[accepted]{icml2023}

\usepackage{amsmath}
\usepackage{amssymb}
\usepackage{mathtools}
\usepackage{amsthm}

\usepackage[inline]{enumitem}
\usepackage{dirtytalk}

\usepackage[capitalize,noabbrev]{cleveref}

\theoremstyle{plain}

\theoremstyle{definition}

\theoremstyle{remark}

\usepackage[textsize=tiny]{todonotes}

\usepackage[capitalize,noabbrev]{cleveref}
\usepackage[inkscapearea=page]{svg}

\definecolor{orange}{rgb}{1,0.5,0}

\icmltitlerunning{The Extractive-Abstractive Axis}

\begin{document}

\twocolumn[
\icmltitle{The Extractive-Abstractive Axis: \\ Measuring Content ``Borrowing" in Generative Language Models}



\icmlsetsymbol{equal}{*}

\begin{icmlauthorlist}
\icmlauthor{Nedelina Teneva}{yyy} 
\end{icmlauthorlist}

\icmlaffiliation{yyy}{Megagon Labs, Mountain View, CA, USA}

\icmlcorrespondingauthor{Nedelina Teneva}{nedteneva@gmail.com}

\icmlkeywords{Machine Learning, ICML}

\vskip 0.3in
]



\printAffiliationsAndNotice{}  

\begin{abstract}
Generative language models produce highly \textit{abstractive} outputs by design, in contrast to \textit{extractive} responses in search engines. Given this characteristic of LLMs and the resulting implications for content Licensing \& Attribution, we propose the the so-called \textit{Extractive-Abstractive axis} for benchmarking generative models and highlight the need for developing corresponding metrics, datasets and annotation guidelines. We limit our discussion to the text modality. 
\looseness=-1
\end{abstract}
\section {Introduction}

The widespread adoption of Large Language Models (LLMs) has created many practical data governance challenges, among which Licensing \& Attribution has emerged  as a key one \cite {jernite2022data}. The interplay between generative language models and copyright law, the fair use doctrine and licensing requirements is of broad research and practical interest to legal practitioners, and increasingly, developers and users of LLMs. This topic is not new: content owners' rights have been of interest to the legal community since the inception of the web and the subsequent wide spread use of search engines \cite{travis2008opting}. Traditionally, search engines have been powered by information retrieval techniques, which take as input a user query and output a query answer by parsing out relevant paragraphs, sentences or phrases from a web-scale corpus of documents to produce an \textit{attributable extractive answer} to the query.

The advent of LLMs -- which \citet{liu2023evaluating} call \say{generative search engines} -- is leading to a paradigm shift from \textit{attributable extractive} question answering and summarization methodologies to increasingly \textit{abstractive} ones. To produce these abstractive responses, generative models \cite{bart, t5} synthesize information from multiple sources/text documents using sequence-to-sequence LLMs such that the generated answers may be highly abstractive or otherwise not readily attributable -- as they are in search engines -- to a specific content source such as a document on the web with a unique URI identifier \footnote{https://datatracker.ietf.org/doc/html/rfc3986} . Given this shift, we propose the \textit{Extractive--Abstractive axis} for quantifying the propensity of LLMs for content borrowing.
We highlight the need for relevant metrics, benchmarks and annotations and list some practical challenges in Section \ref{challenges}.

\vspace{-8pt}
\section {The Extractive--Abstractive Axis}
Being able to quantify a generative language model's extractiveness/abstractiveness level -- in other words where the model lies on what we call the \textit{Extractive-Abstractive axis} -- with respect to one or several sources (e.g., a text snippet, a web page or social media post), is necessary for evaluating whether (and how much) a generative AI application is using content from copyrighted or licensed sources.  Intuitively, LLM answers with high levels of content borrowing in the absence of proper attribution constitute a higher risk for copyright infringement. By way of a practical example: a news publisher would like to determine if their article was used for training a LLM without their permission. 
If the publisher had access to the LLM pre-training and fine tuning corpus they can examine each training document and compare it to their article. If the LLM is only commercially accessible through APIs (e.g., ChatGPT \cite{bubeck2023sparks}), the publisher may query it in an attempt to examine if its responses contain snippets from their article. Depending on the abstractiveness level of the responses, the publisher may be facing a rather complex prompting task while at the same time providing additional training information to an already potentially copyright-infringing LLM operator.

Quantification along the Extractive-Abstractive axis is of practical use to content owners, the developers of generative language applications and third parties for several reasons: 
\begin{enumerate*}[label=\textbf{(\arabic*)}]
\item Being able to quantify the level of language/content borrowing will allows content owners or third parties such as   algorithmic auditors \cite{raji2019actionable} to quantify how prone a trained LLM  is to content borrowing. 
\item Such metrics will enable the designers of generative language-based applications to minimize their legal risks (e.g. copyright infringing)  by identifying highly extractive responses at inference/run time.
\item More generally, such tools  will also help organizations with the assessment of LLM related Licensing \& Attribution risks pre- and post-deployment or liability assessment of off-the-shelf tools such as black box LLMs.
\item In courts these metrics can potentially be used for quantifying if a LLM-generated text is substantially similar to copyrighted content (or derivative of such). 
We may even imagine cases in which the empirical  propensity (measured on benchmarking datasets) of generative models for borrowing large amounts of content may also play a role in Licensing \&  Attribution matters.

\end{enumerate*}
\vspace{-8pt}
\section{Metrics, Datasets and Annotation Tasks}\label{metrics}
\textbf{Metrics}. Existing Natural Language Processing (NLP) tasks such as question answering, machine translation, extractive and abstractive summarization (see Appx. \ref{nlptasks} for definitions) use various automatic  metrics 
to measure the similarity between generated answers and the true \say{gold} answers. Some of these include (see  \citet{liu2023towards, fabbri2021summeval} for additional ones): 
\textbf{(1)} token overlap metrics (e.g., ROUGE \cite{lin2004rouge} and BLEU \cite{papineni2002bleu}) compare the similarity between two texts based on the $n$-grams (contiguous sequence of tokens) overlap between them; 
\textbf{(2)} vector-based metrics such as BERTScore \cite{zhang2019bertscore} and BARTScore \cite{yuan2021bartscore} measure text sequence similarity based on text representations learnt by neural models; 
\textbf{(3)} metrics relying on the assumption that if two texts are similar, they should be able to address the same set of questions -- one example is QAEval \cite{qaeval}; 
\textbf{(4)} additional metrics such as $n$-grams ratio \cite{narayan2018don} or coverage \cite{newsroom} are used for measuring summarization quality.

In principle, while some of the above mentioned NLP metrics can be repurposed to measure LLMs along the Extractive-Abstractive axis in matters related to Licensing \& Attribution, no empirical studies exist on this topic and there are no evaluation benchmarks to guide such analysis. Since these automatic metrics have not been previously applied in the context of Licensing\& Attribution, there are no empirical studies of whether  they correlate with content owners'  perception of Licensing \& Attribution.

\textbf{Datasets and Human Annotations}.
Like all NLP models, LLMs are evaluated with respect to the downstream \textit{user perception} of the answer quality -- see \citet  {rogers2023qa} for a review and taxonomy of the vast number of NLP datasets.
They are, however, not evaluated with respect to the \textit{content owners' perception} of how well their content is used for answering users' questions. Given LLM's propensity for content borrowing,  it is critical that the experience and rights of content owners are balanced with those of application users.

A simple approach for benchmarking content owners' perception of Licensing \& Attribution quality is by re-purposing existing NLP datasets.  For example, summarization tasks, which already rely heavily on human annotation, may be particularly well suited for benchmarking generative models along the Extractive-Abstractive axis.
\begin{figure}[t]
\vspace{-8pt}
\begin{center}
\includegraphics[width=\columnwidth]{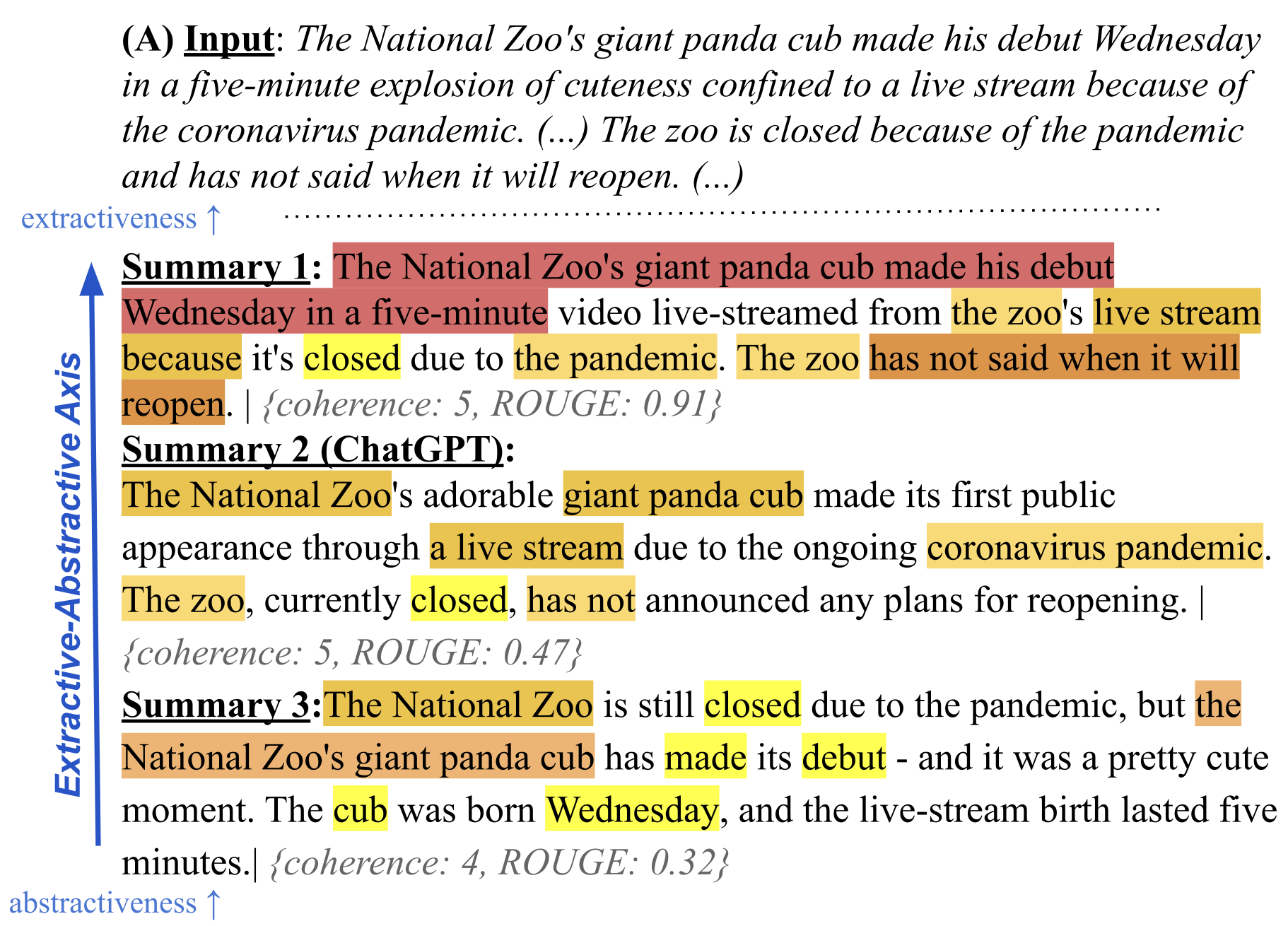}
\includegraphics[width=0.85\columnwidth]{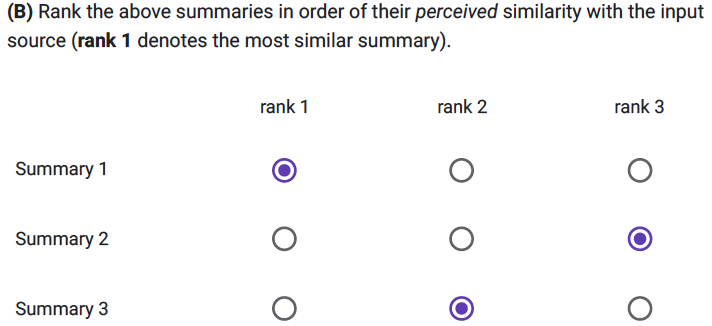}\vspace{-6pt}
\caption{\textbf{(A)} Illustration of $3$ summaries (of the same input) along the Extractive-Abstractive Axis (similarity measured by ROUGE) and their example coherence scores. Text by \citet{dreyer-tradeoff}, with permission. 
\textbf{(B)} Example guideline for annotating the perceived similarity between the summaries.
Details in Appx. \ref{figdetails}.}
\label{extrabstr}
\end{center}
\end{figure}
Currently, summarization tasks are benchmarked using human annotators who rate the generated summaries along dimensions such as summary
 relevance, fluency, coherence \cite{fabbri2020summeval} or the level of factual alignment between the summary and the underlying source text being summarized (e.g., answer  consistency \cite{fabbri2020summeval}, faithfulness \cite{ladhak2021faithful} and  factuality \cite {dreyer-tradeoff}).
As an example, Figure \ref{extrabstr}A shows summaries with their example coherence scores.
These summarization benchmarks can be augmented with (legal) expert annotation of Licensing \& Attribution quality assessing whether, e.g. (1) the similarity between the input text and summary is acceptable, (2) there are copyright concerns, (3) any extractive snippets are properly attributed to the source.
Some of these dimensions may be more easy to assess in a comparative manner, rather than individually as long as proper annotator agreement is established. Figure \ref{extrabstr}B illustrates this point with an example annotation question which ranks the 3 summaries in order of decreasing perceived similarity.

\section{Practical Challenges and Limitations}\label{challenges}

There are several practical challenges associated with measuring generative models along the Extractive-Abstractive axis which we categorize below.

\textbf {Evaluation Challenges}: Human evaluation, especially of longer answers, is a hard and actively studied research problem \cite{rogers2023qa}. Content Licensing \& Attribution nuances and expertise required can pose challenges to the human evaluation of the generated responses described in Section \ref{metrics} and Figure \ref{extrabstr}. Additionally, while this study focuses mainly on English language and we note that  content borrowing may be different in other languages.

\textbf{Usability Challenges and Conflicting Interests}: 
(1) Correlation between faithfulness/factuality and extractiveness  \cite {dreyer-tradeoff,ladhak2021faithful} observed in summarization tasks implies that a certain level of extractiveness may be needed in the generated answers in order to balance mis/disinformation concerns. This observation may heighten the necessity of measuring content borrowing in LLMs along the Extractive-Abstractive axis. 
(2) Interactions with LLMs can be used as additional training signals for the underlying LLM system so a practical challenge is how generative models can be audited for Licensing \& Attribution purposes without further aiding their development.

\textbf{Ethical Challenges}: Content borrowing poses numerous ethical challenges in addition to legal ones. In order to mitigate such challenges, incentives and broader policies may be needed in order to alleviate the concerns of both content owners and LLM end users. Adversarial scenarios also needs to be considered: LLMs can be  tuned for specific abstraction levels which means that copyrighted content used for pre-training and fine tuning can be intentionally obfuscated. In such cases, developing methodologies for identifying copyright infringement in black box LLMs becomes even more critical. 

\section*{Acknowledgements}
We would like to thank John Santerre for their input in the early stages of the project; Estevam Hruschka, Sajjadur Rahman and the anonymous reviewers for their draft feedback and suggestions. 


\bibliography{paper}
\bibliographystyle{icml2023}

\newpage
\appendix
\onecolumn
\section{Appendix}\label{appx}

\subsection{Figure Details}\label{figdetails}
The figure shows three summaries of the input text snippet shown at the top of the figure; each summary is of different degree of extractives/abstractiveness (measured by ROUGE score \cite{lin2004rouge}). Summaries $1$ and $3$ are reproduced with permission from \citet{dreyer-tradeoff}'s study. Summary $2$ was obtained by prompting ChatGPT (using the free plan) in May 2023 as follows: \say{Summarize this snippet: \say{The National Zoo's giant panda cub made his debut Wednesday in a five-minute explosion of cuteness confined to a live stream because of the coronavirus pandemic. The zoo is closed because of the pandemic and has not said when it will reopen.}}

For each summary in (A), we show 1) an example scoring of the coherence; and 2) automatically computed ROUGE-L score (recall) between the summary and the input using the \texttt{py-rouge} library (\href{https://pypi.org/project/py-rouge/}{https://pypi.org/project/py-rouge/}). 
Following  \citet{dreyer-tradeoff}, fragments extracted from the input are marked from red (longer fragments) to yellow (shorter fragments). 

\subsection{NLP Task Definitions}\label{nlptasks}
\textbf{Question Answering} refers to the task of answering asked by humans in natural language using either a pre-structured database or a collection of text documents (see \cite{soares2020literature} for a review). There are various  subtypes of question answering such as factoid question answering or  multiple choice question answering.

\textbf{Summarization} tasks aim produce summaries of single or multiple documents to answer questions that require longer responses. The goal is to convey the key information in the input text. In \textit{extractive} summarization, the summarizers identify the most important sentences in the input, which can be either a single document or a cluster of related documents, and string them together to form a summary \cite{nenkova2012survey}. In \textit{abstractive} summarization, the summary contains synthesized text which may not be explicitly present in the input text.

\end{document}